\documentclass[conference]{IEEEtran}
\IEEEoverridecommandlockouts
\usepackage{cite}
\usepackage{amsmath,amssymb,amsfonts}
\usepackage{algorithmic}
\usepackage{graphicx}
\usepackage{subfigure}
\usepackage{multirow}
\usepackage{textcomp}
\usepackage{xcolor}
\usepackage{hyperref}
\usepackage{adjustbox}

\def\BibTeX{{\rm B\kern-.05em{\sc i\kern-.025em b}\kern-.08em
    T\kern-.1667em\lower.7ex\hbox{E}\kern-.125emX}}
\begin{document}

\title{Exploiting Richness of Learned Compressed Representation of Images for Semantic Segmentation\\}

\author{\IEEEauthorblockN{Ravi Kakaiya, Rakshith Sathish, Debdoot Sheet}
\IEEEauthorblockA{\textit{Indian Institute of Technology Kharagpur}\\
West Bengal, India}
\and
\IEEEauthorblockN{Ramanathan Sethuraman}
\IEEEauthorblockA{\textit{Intel Technology India Pvt. Ltd.}\\
Bengaluru, India}
}

\maketitle

\begin{abstract}
Autonomous vehicles and Advanced Driving Assistance Systems (ADAS) have the potential to radically change the way we travel. Many of such vehicles currently rely on segmentation and object detection algorithms to detect and track objects around its surrounding. The data collected from the vehicles are often sent to cloud servers to facilitate continual/life-long learning of these algorithms. Considering the bandwidth constraints, the data is compressed before sending it to servers, where it is typically decompressed for training and analysis. In this work, we propose the use of a learning-based compression Codec to reduce the overhead in latency incurred for the decompression operation in the standard pipeline. We demonstrate that the learned compressed representation can also be used to perform tasks like semantic segmentation in addition to decompression to obtain the images. We experimentally validate the proposed pipeline on the Cityscapes dataset, where we achieve a compression factor up to $66 \times$ while preserving the information required to perform segmentation with a dice coefficient of $0.84$ as compared to $0.88$ achieved using decompressed images while reducing the overall  compute by $11\%$.
\end{abstract}


\section{Introduction}
Autonomous driving and Advanced Driving Assistance Systems (ADAS) have the potential to revolutionize the way we commute. With the rise of self-driving cars, we can expect reduced traffic congestion, improved safety, and greater efficiency in transportation. One of the key tasks for self-driving cars is segmentation \cite{b1} and object detection \cite{b2}, which involves detecting and tracking objects in the vehicle's surroundings, such as other cars, pedestrians, and traffic signs. This requires advanced algorithms, such as deep neural networks, which demand significant computational resources. Moreover, the data collected by autonomous vehicles during operation is vast, and processing it in real-time is a significant challenge. The data collected by autonomous vehicles are often sent to cloud servers \cite{b3} for additional post-processing and tasks like fine-tuning the model of continual learning to address this challenge. However, sending large amounts of data to the cloud presents challenges, including data security, latency, and bandwidth constraints. In this paper, we discuss how learning-based compression codecs can potentially solve these challenges while optimizing the computational requirements of the data processing pipeline involved in sending and processing enormous amounts of data in real-time.
One of the biggest challenges associated with compressing images using deep learning-based compression engines is balancing the trade-off between compression efficiency and the preservation of important information. The compressor model must compress the images while retaining sufficient information for accurate analysis and safe operation of the vehicle. Additionally, the complexity of the deep learning algorithms used for compression can lead to high computational costs and increased latency. The network architecture must be designed with minimal compute cost to overcome this.

\begin{figure}[t]
\centering
  \includegraphics[width=\linewidth]{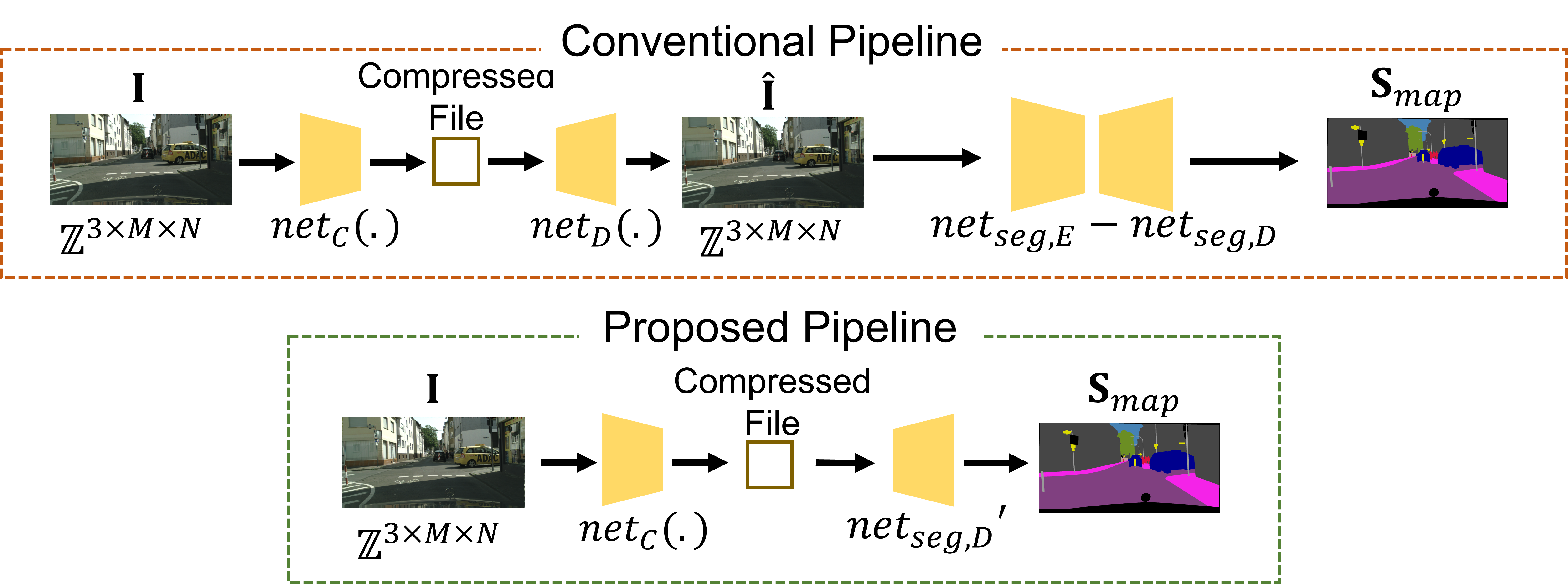}%
  \caption{Overview of the proposed method compared with the conventional segmentation pipeline using compressed images. Here, $net_C$ denotes the compressor of the learning-based compression engine, $net_D$ denotes the corresponding decompressor. The encoder and decoder blocks of the segmentation network are denoted   using $net_{seg,E}$ and $net_{seg,D}$, respectively. $\mathbf{I}$ and $\hat{\mathbf{I}}$ represents the original and decompressed images. $\mathbf{S}_{map}$ represents the generated segmentation map}
  \label{fig:graph_abstract}
\end{figure}

\noindent\textbf{Prior art:} Learning-based compression engines have shown enormous potential in compressing high-resolution medical images at high compression factors (CF) \cite{b4}. The effectiveness of neural architecture search in balancing the trade-off between compression efficiency and information preservation while minimizing computational costs is demonstrated in \cite{b5}. However, the cross-domain adaptability of these design principles is yet untested. A significant portion of the total computational cost incurred during decompression can be avoided if image analysis tasks like segmentation can be performed in the compressed domain. A joint learning framework that learns to predict labels and reconstruct the image from a compressed latent vector is proposed in \cite{b6}. However, this framework is not scalable as the tasks are learned together. If the framework needs to be extended for a new application task on the same data, for example, object detection, then the entire framework must be retrained from start. Compression of histopathology images to produce a compressed representation, which can be used to generate image-level label, is proposed in \cite{b7}.
\newline
\noindent\textbf{Our approach:} Inspired by the success of \cite{b7} in performing classification using compressed representation, we propose a method for segmentation using compressed representations directly. In this paper, we extend the design principles proposed in \cite{b5} to learn the compressed representations and evaluate their generalisability and cross-domain adaptability. In our approach, the compressed representation obtained from the compressor is provided as input to the semantic segmentation network adopted from ~\cite{b9}. Through this, we experimentally demonstrate that the compressed representation learned with the objective of minimizing which are image reconstruction error preserves features rich enough to perform image analysis tasks like segmentation.

\section{Method}
\subsection{Compression of High-Resolution Driving Images Sequences}
\begin{figure}[!h]
\centering
  \includegraphics[width=0.8\linewidth]{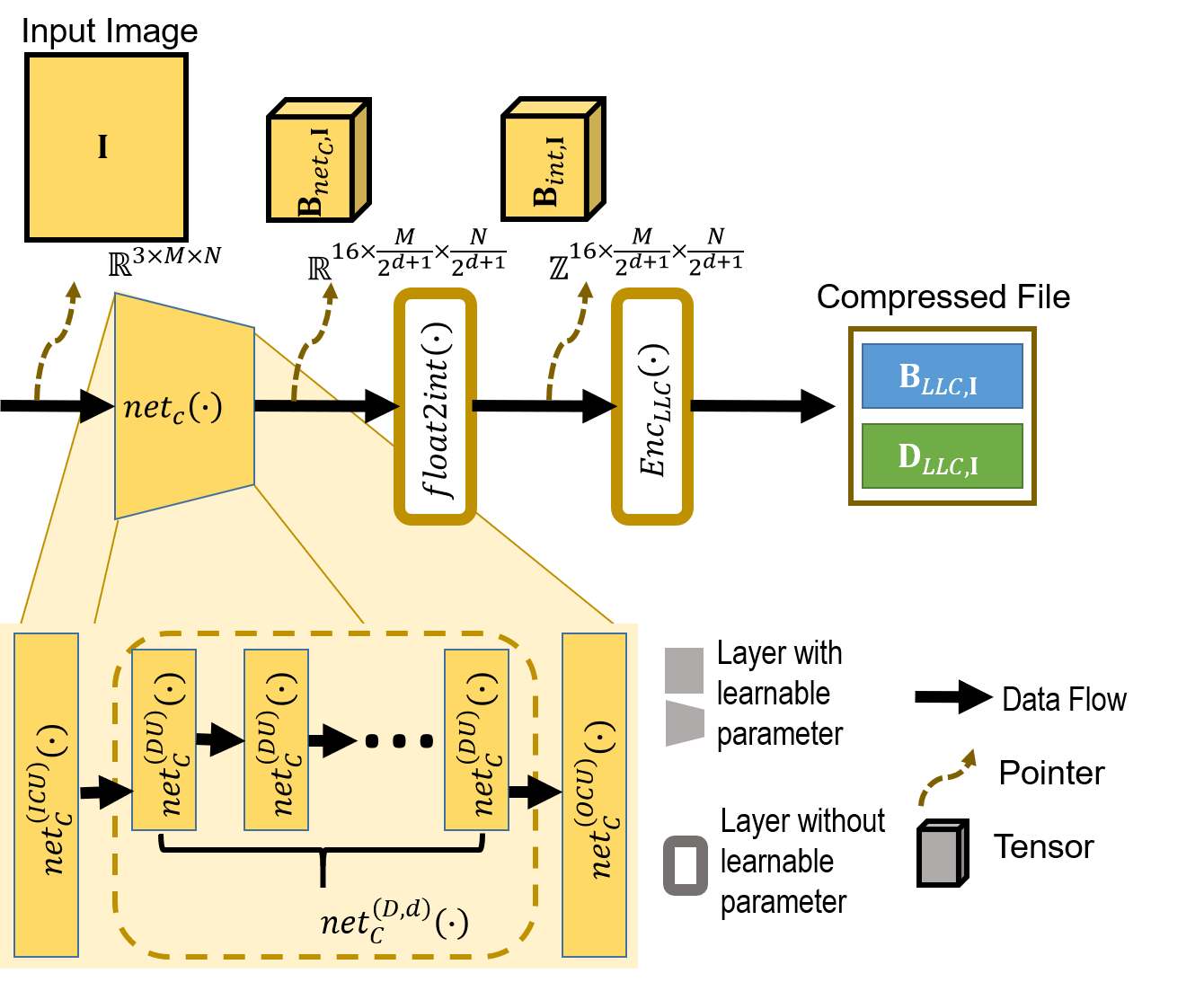}%
  \caption{Compression pipeline \cite{b5}}
  \label{fig:netc}
\end{figure}

\noindent\textbf{Compressor:} An image from high resolution driving video sequence $\mathbf{I} \in \mathbb{Z}^{3 \times M \times N}$ is passed through a convolutional compressor ($net_{C}(.)$), to obtain the output tensor $\mathbf{B}_{net_{C}}$. This output is then converted to an integer tensor $\mathbf{B}_{int}$ using $float2int(\cdot)$ operation to lower bit length representation. A lossless compression~\cite{b8} is then performed on $\mathbf{B}_{int}$ to obtain the learned compressed representation $\mathbf{B}_{LLC}$ and the encoding dictionary $\mathbf{D}_{LLC}$. The compressor is illustrated in Fig.~\ref{fig:netc}.
\begin{equation}\label{comp}
    [\mathbf{B}_{LLC},\mathbf{D}_{LLC}] = Enc_{LLC}(float2int(net_{C}(\mathbf{I})))
\end{equation}
\begin{equation}\label{f2i}
    float2int(x,n) = (2^{n}-1)\frac{x-\delta }{\Delta -\delta}
    \end{equation}
    where, $x \in \mathbf{B}_{net_{c}}$ ,$\Delta = max(\mathbf{B}_{net_{c},\tau}), \delta = min(\mathbf{B}_{net_{c},\tau})$. $n$ is the bit length used to store the unique values in the $\mathbf{B}_{int}$.

%\FloatBarriers
\begin{figure}[!h]
\centering
  \includegraphics[width=0.8\linewidth]{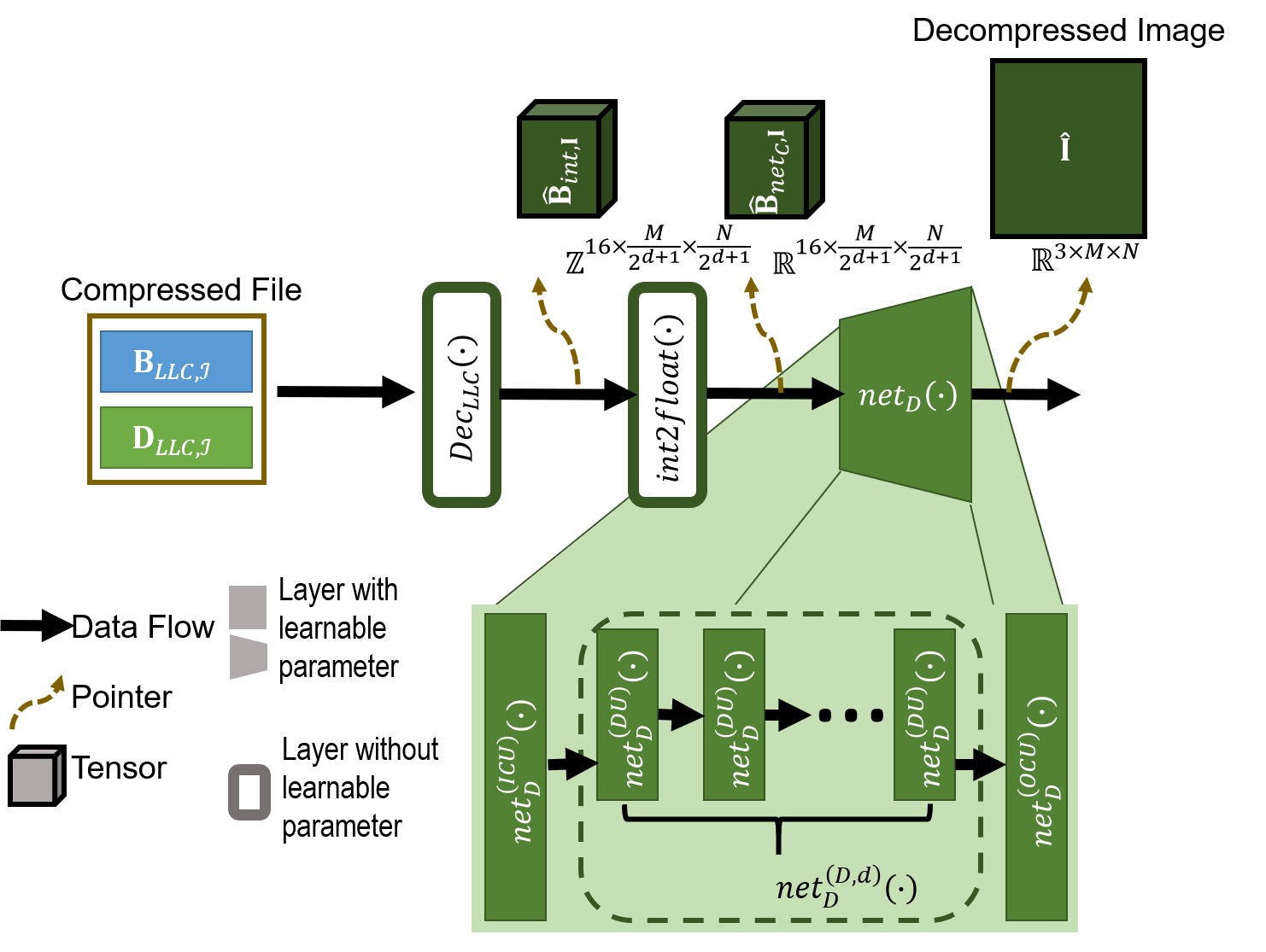}%
  \caption{Decompression pipeline \cite{b5}}
  \label{fig:netd}
\end{figure}
%\FloatBarriers

\noindent\textbf{Decompressor:} The learned compressed representation $\mathbf{B}_{LLC}$ and the encoding dictionary $\mathbf{D}_{LLC}$ are converted to an integer tensor $\hat{\mathbf{B}}_{int}$ by the lossless decompressor~\cite{b8}. Floating point representation $\mathbf{B}_{net_{D}}$ using obtained by performing $int2float(\cdot)$ operation on $\hat{\mathbf{B}}_{int}$. This intermediate representation $\mathbf{B}_{net_{D}}$ is then passed through $net_{D} (.)$ to obtain the decompressed image $\hat{\mathbf{I}}$. The decompressor is illustrated in Fig.~\ref{fig:netd}.

\begin{equation}\label{decom}
    \hat{\mathbf{I}} = net_{D}(int2float(Dec_{LLC}(\mathbf{B}_{LLC},\mathbf{D}_{LLC})))
\end{equation} 
\begin{equation}\label{inflt}
        int2float(x,n) =\frac{x}{2^n-1}(\Delta-\delta)+\delta
        \end{equation}
        where, $x \in \hat{\mathbf{B}_{int}}$ and $n$ is the bitlength.

\noindent\textbf{Training routine:} During training, the weights are updated for both $net_{C}(.)$ and $net_{D}(.)$ with respect to gradients calculated using the reconstruction error between $\mathbf{I}$ and $\hat{\mathbf{I}}$.

\noindent\textbf{Inference routine:} During inference, we compress the image $\mathbf{I}$ using $net_{C}(.)$ to generate its corresponding compressed representation $\mathbf{B}_{net_C}$. These representations are saved and used for training $net_{seg, D'}$. It is to be noted that $net_{D}(.)$ is not required here, which results in lowered computation cost of the pipeline.

\subsection{Segmentation}

We use the dual graph convolutional neural network (DGCN) architecture proposed by \cite{b9} to perform segmentation. The segmentation network $net_{seg}(\cdot)$ consists of a backbone network that provides a feature map $\mathbf{X}$ and dual graph convolutional layers, which effectively and efficiently models contextual information for semantic segmentation. We use ResNet-$50$~\cite{b10} architecture as our backbone network, which consists of $5$ residual blocks, similar to \cite{b9}.

\begin{figure}[!h]
\centering
  \includegraphics[width=0.9\linewidth]{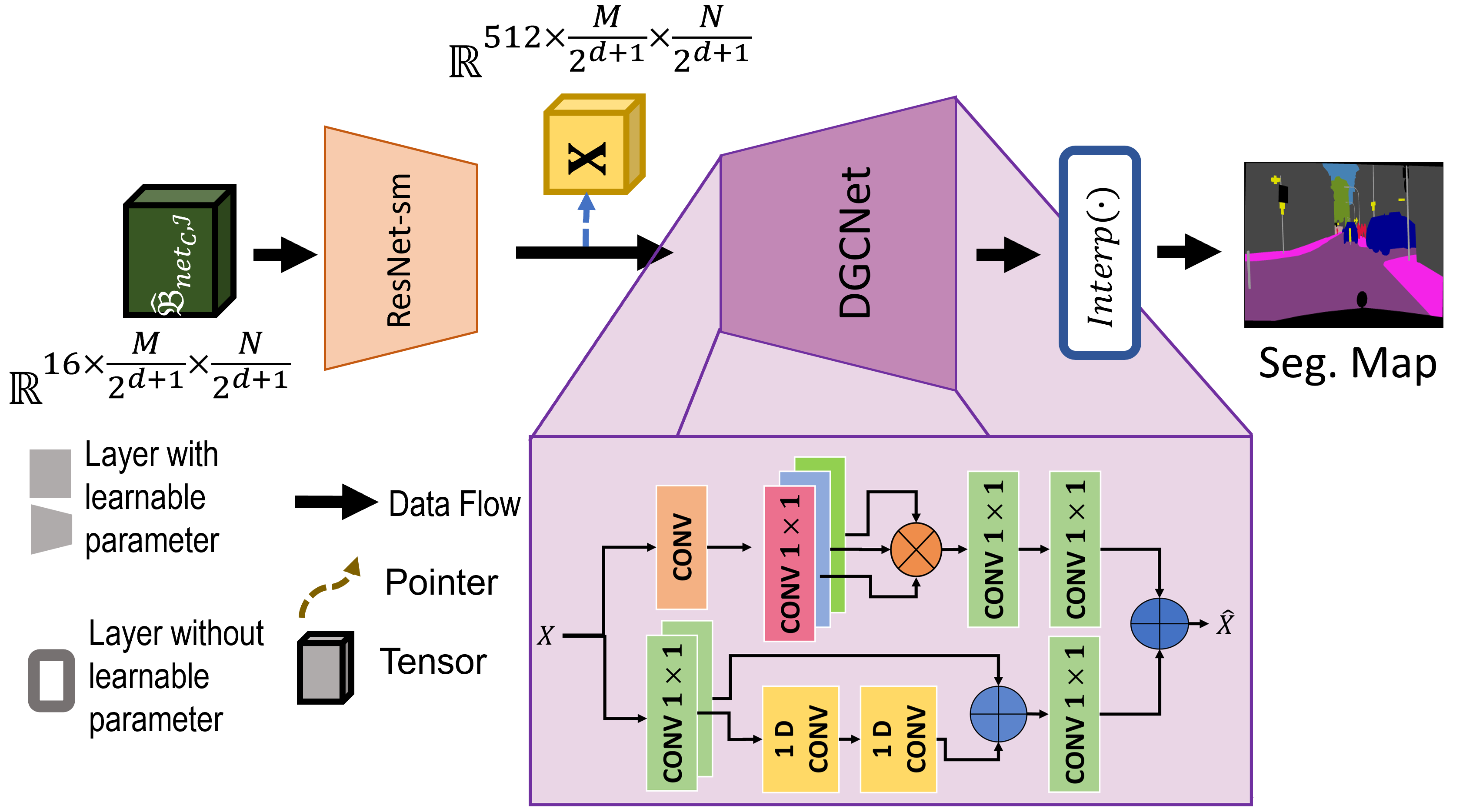}%
  \caption{Architecture of segmentation network $net_{seg,D'}$ denoted in Fig \ref{fig:graph_abstract}}
  \label{fig:net_segd}
\end{figure}

In order to perform segmentation on the compressed representations $\mathbf{B}_{net_C} \in \mathbb{R}^{16\times \frac{M}{d+1}\times \frac{N}{d+1}}$, we modify the original architecture of $net_{seg}(\cdot)$ by replacing the backbone with a smaller modified version of the ResNet-$50$ network, referred further to as ResNet-sm. ResNet-sm has an initial convolutional layer with $16$ input channels and $2$ residual blocks. The output of ResNet-sm is provided as input to $DGCNet(\cdot)$ to obtain the segmentation predictions. The overall architecture of the modified segmentation network $net_{seg, D'}(\cdot)$ is shown in Fig.~\ref{fig:net_segd}.

%\FloatBarriers

\section{Experiments}

\noindent\textbf{Dataset description:} Cityscapes \cite{b11} dataset contains $5,000$ images of size $1,024 \times 2,048$  with polygon annotations for 34 classes. We use the validation set provided in the dataset as our held-out test set, and the training set is divided into our training and validation sets in an $80:20$ ratio. The compression model ($net_{C}(\cdot) - net_{D}(\cdot)$) for all baselines and the proposed method are trained with patches of size $256 \times 256$, and segmentation models ($net_{seg}(\cdot)$,$net_{seg, D^{'}}(\cdot)$) were trained by using non-overlapping   patches of size $840 \times 840$, respectively, which were extracted from the training set without overlapping.

% \footnotetext[1]{\url{https://cityscapes-dataset.com/}}
% \footnotetext[2]{\url{https://github.com/mcordts/cityscapesScripts}}

\noindent\textbf{Model parameters:}
The compression model ($net_{C}(\cdot) - net_{D}(\cdot)$) was trained for $100$ epochs with Adam as optimizer using a step learning rate scheduler with an initial learning rate of $1 \times 10^{-2}$, step size of $10$ and multiplication factor $\gamma$ of $0.75$. The segmentation decoder ($net_{seg, D^{'}}$) was trained for $40,000$ iterations using SGD as the optimizer with an initial learning rate of $1 \times 10^{-3}$. Mean square error and cross-entropy loss were chosen as loss functions for compression and segmentation, respectively.

\noindent\textbf{Baselines:} \textit{BL $1$}- The segmentation network $(net_{seg}(\cdot)$) is trained and inferred using original images available in the dataset. \textit{BL $2$}- $net_{seg}(\cdot)$ is trained using original images and inferred using decompressed images obtained from $net_{D}(\cdot)$. \textit{BL $3$}- $net_{seg}(\cdot)$ is trained and inferred using decompressed images obtained from $net_{D}$. \textit{BL $4$}- $net_{seg}(\cdot)$ is trained and inferred using \textit{JPEG} images having compression factor of $66$.

\noindent\textbf{Proposed method:} The $net_{seg, D}$ was trained using $4,944$ compressed representations paired with corresponding segmentation maps.

\section{Results and Discussion}

\subsection{Evaluation of Compression} \label{eval_comp}
\begin{figure}[!h]
\centering
  \includegraphics[width=0.95\linewidth]{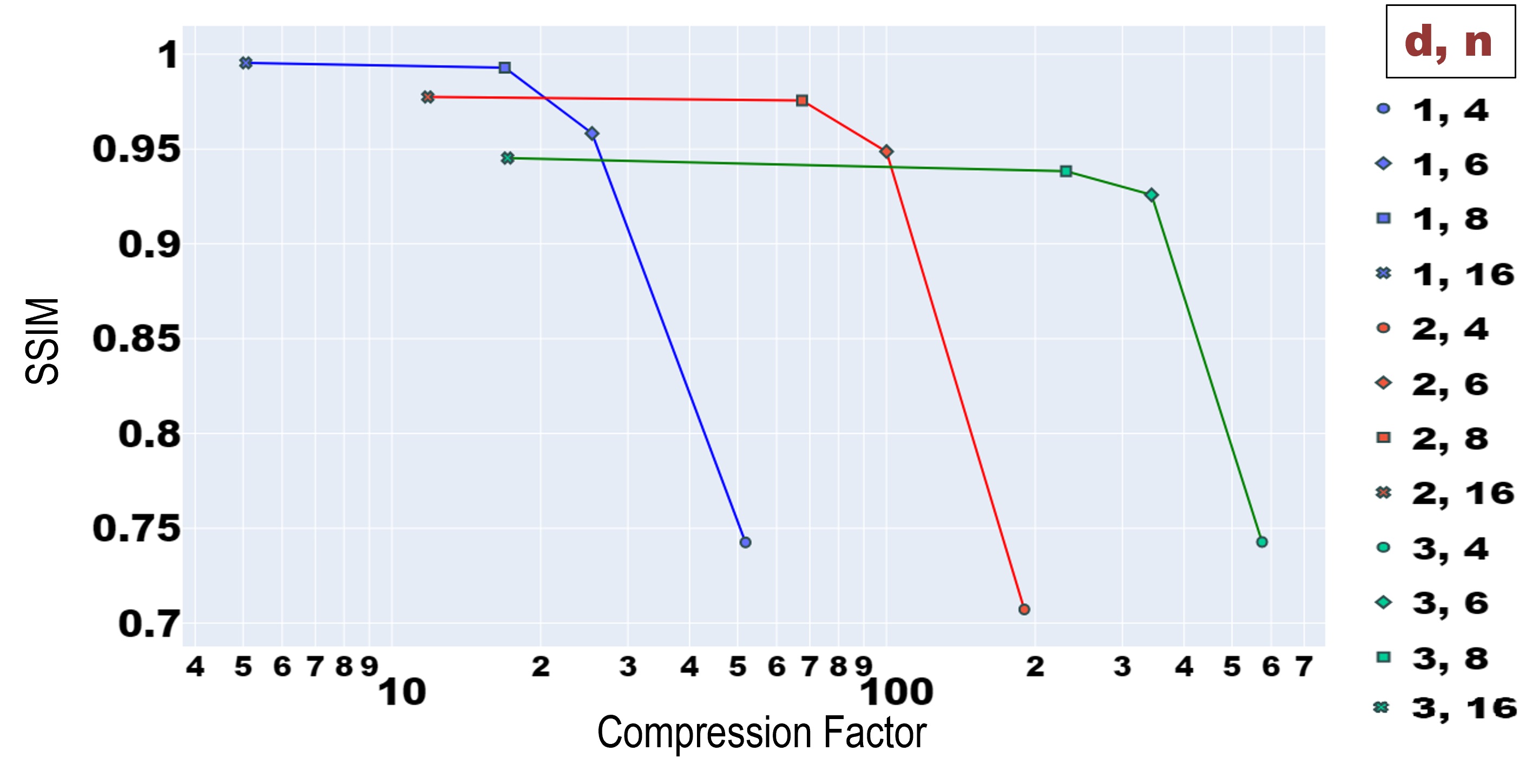}%
  \caption{Quality of compression in terms of SSIM at various compression factors. The change in CF is achieved by varying the network depth ($d$) and bit length ($n$).}
  \label{fig:comp_ssim}
\end{figure}

\begin{figure}[!h]
\centering
  \includegraphics[width=0.95\linewidth]{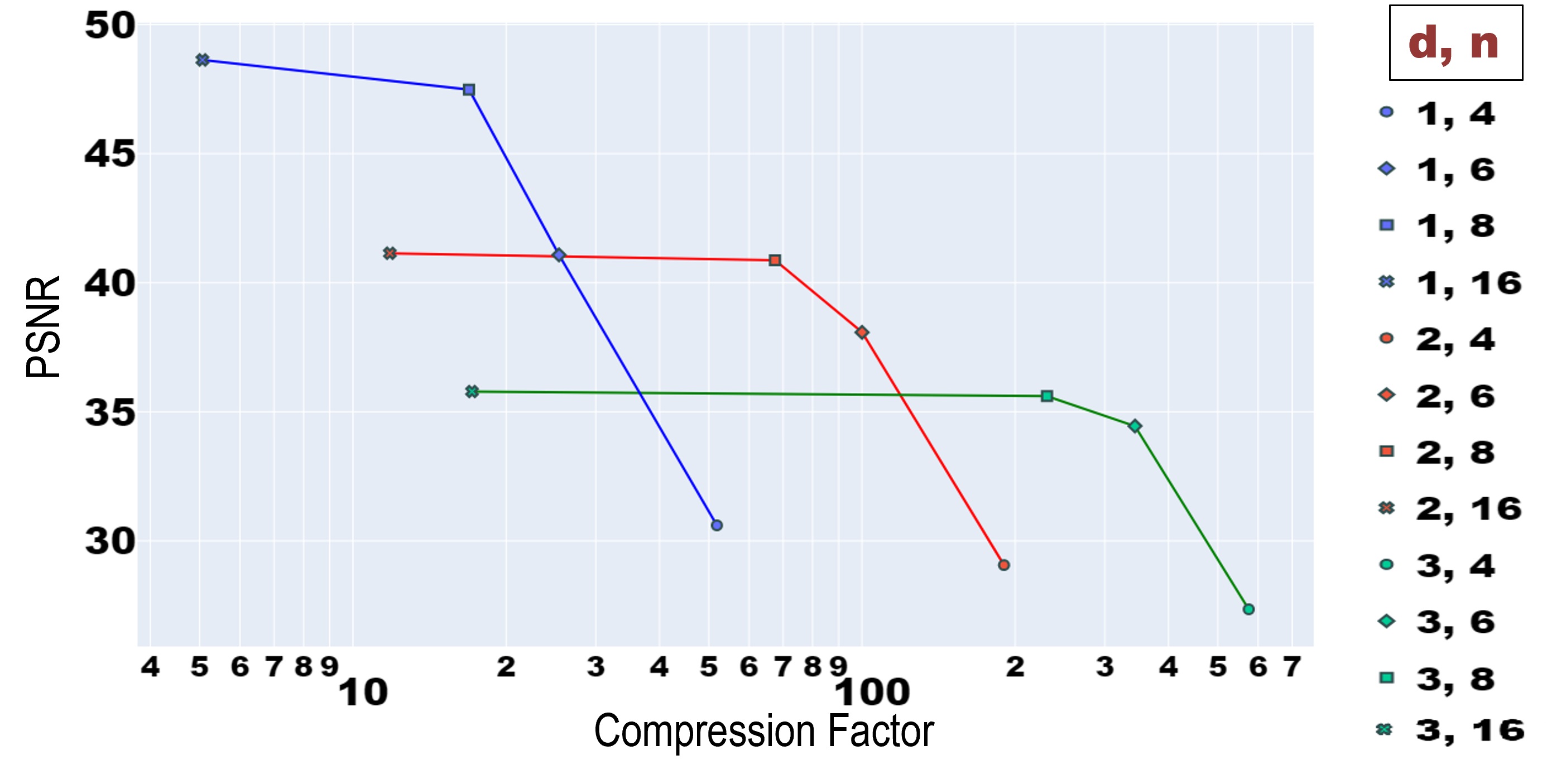}%
  \caption{Quality of compression in terms of pSNR at various compression factors. The change in CF is achieved by varying the network depth ($d$) and bit length ($n$).}
  \label{fig:comp_psnr}
\end{figure}

The quality of compression in terms of SSIM and pSNR at varying network depth or the number of digest units ($d$) and bit length ($n$) is shown in Fig.~\ref{fig:comp_ssim} and Fig.~\ref{fig:comp_psnr}, respectively. It can be observed that for all values of $d$ in range 1 to 3, we do not observe significant degradation in the quality of the decompressed image. However, as shown in \ref{fig:comp_ssim} and \ref{fig:comp_psnr} for values of $n$ less than $6$, we can observe a noticeable drop in performance. 

Further, we can observe that with a learnable compression codec, we can compress the images up to $200 \times$ without a significant drop in performance for a bit length of $8$. These results further corroborate the observations made by \cite{b5} in the case of radiology image compression. Hence, we can safely assume that the design strategy for high-density compressors set forward by ~\cite{b5} can be adopted across domains.

%\FloatBarriers

%\FloatBarriers

\begin{figure*}[t]
\centering
\subfigure[Ground Truth]{\includegraphics[width=0.38\linewidth]{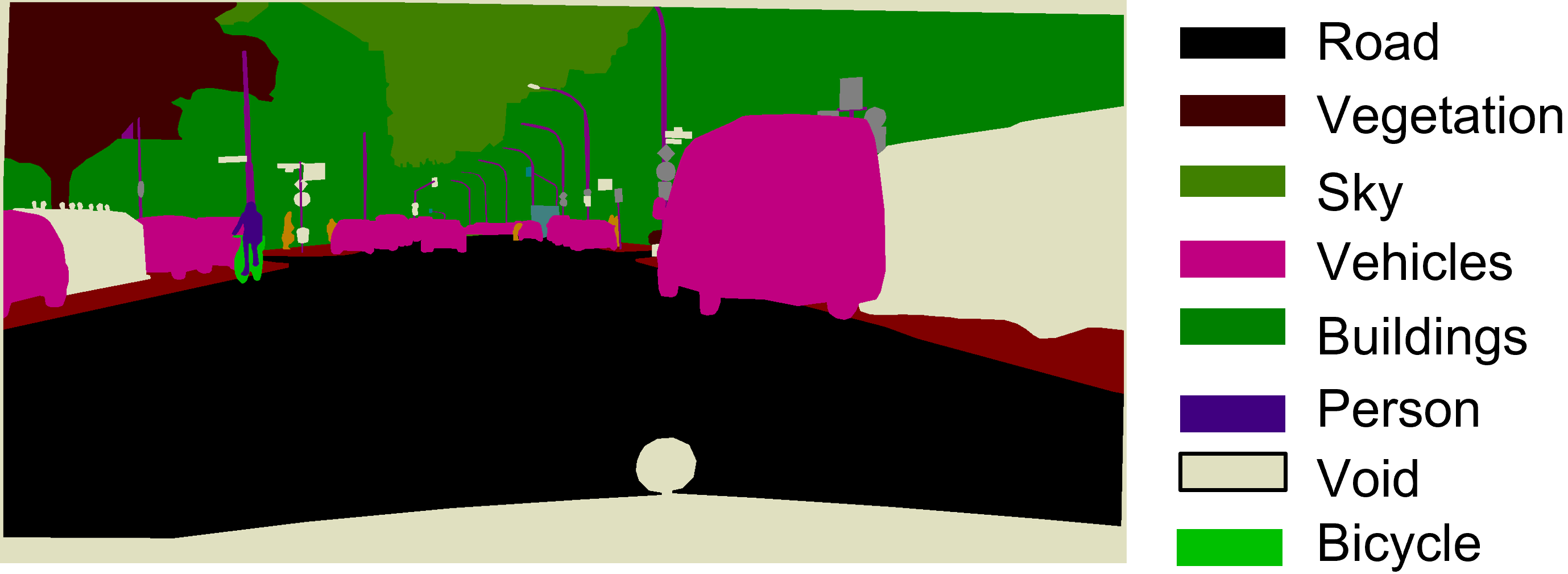}\label{fig:gt}}
\subfigure[$net_{seg,D}$, Dice = $0.86$]{ \includegraphics[width=0.28\linewidth]{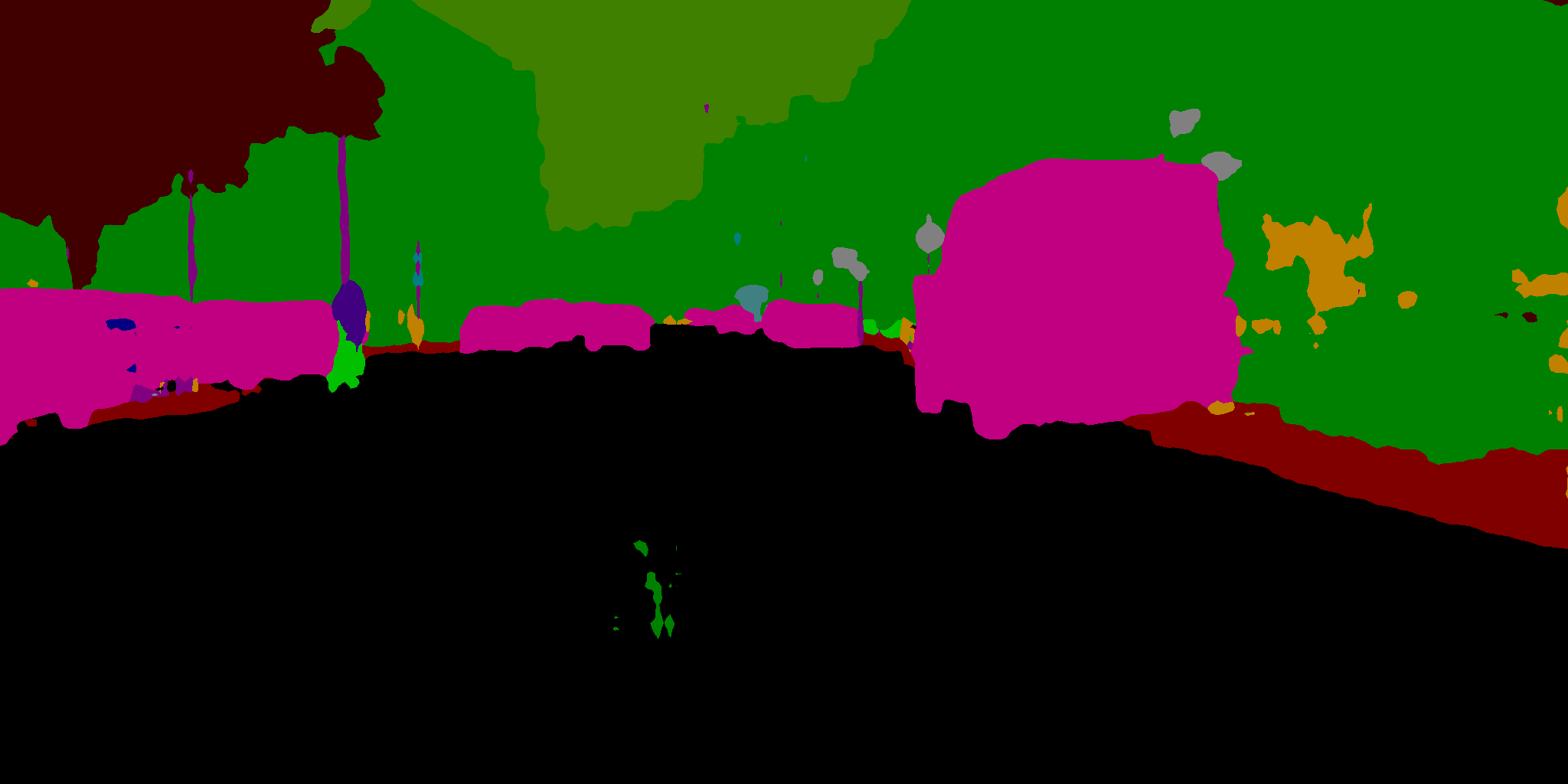}\label{fig:e1}}
\subfigure[\textit{BL 1}, Dice = $0.89$]{\includegraphics[width=0.28\linewidth]{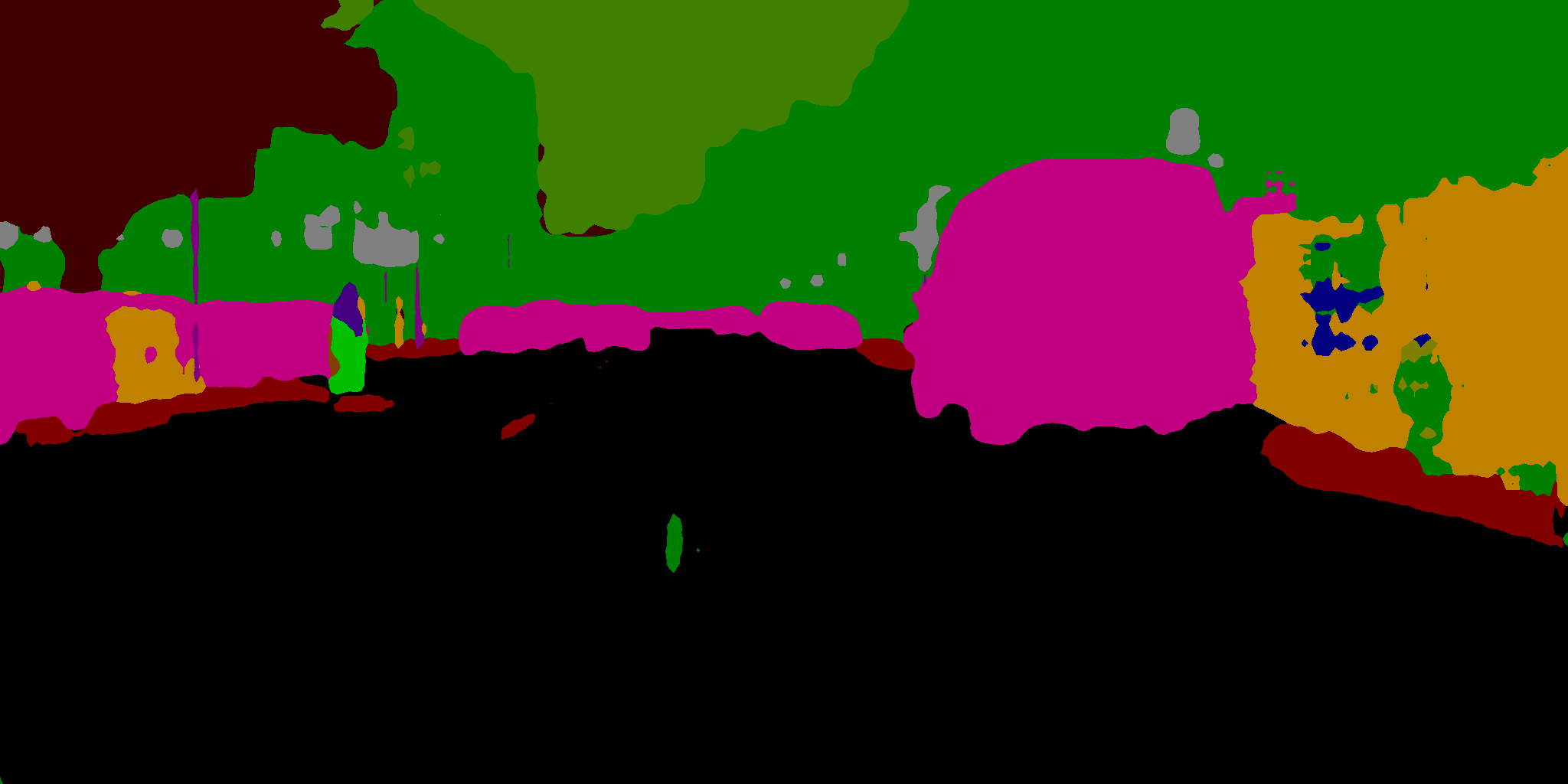}\label{fig:e2}}

\subfigure[\textit{BL 2}, Dice = $0.86$]{\includegraphics[width=0.28\linewidth]{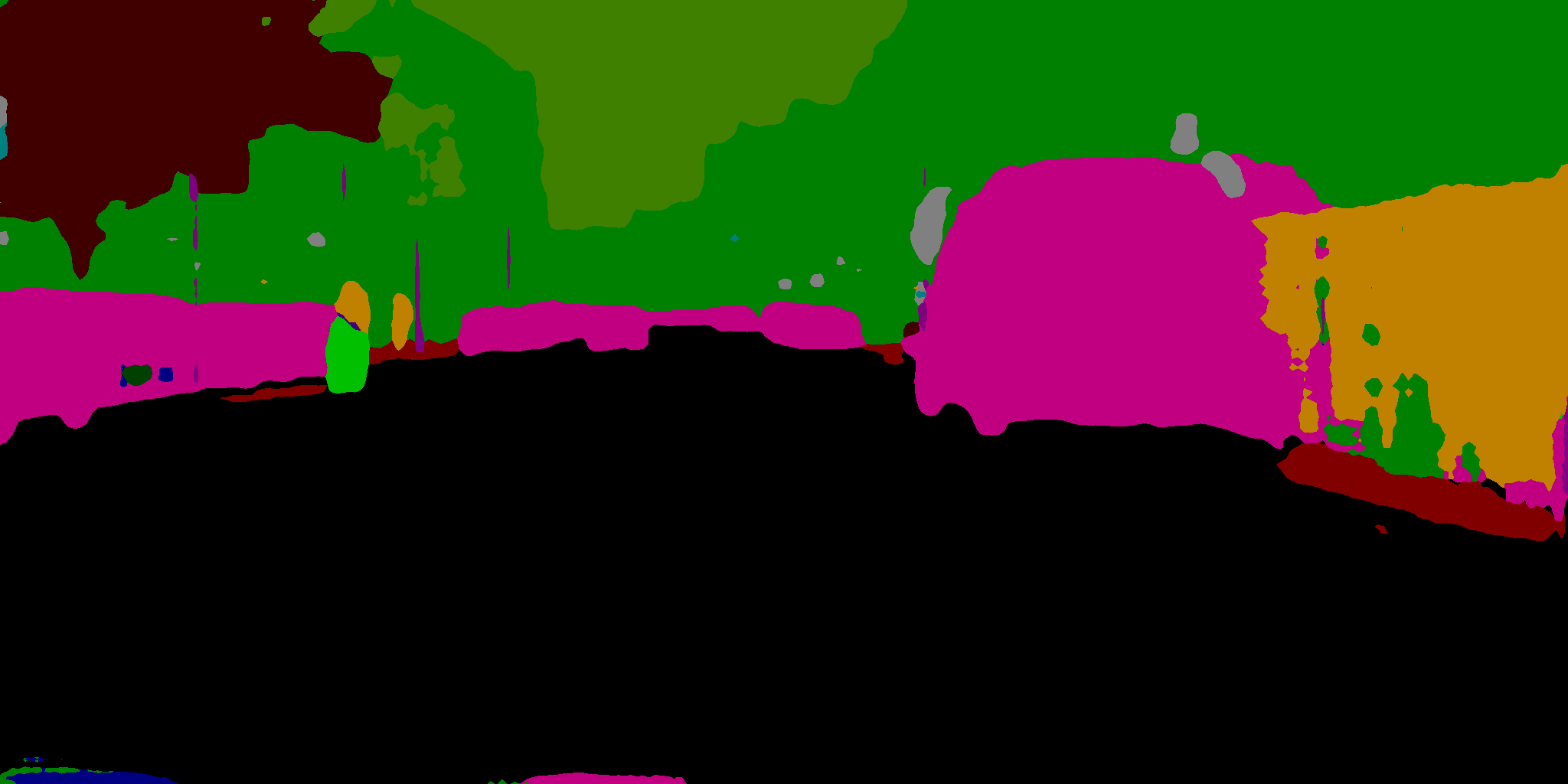}\label{fig:e4}}
\subfigure[\textit{BL 3}, Dice = $0.87$]{ \includegraphics[width=0.28\linewidth]{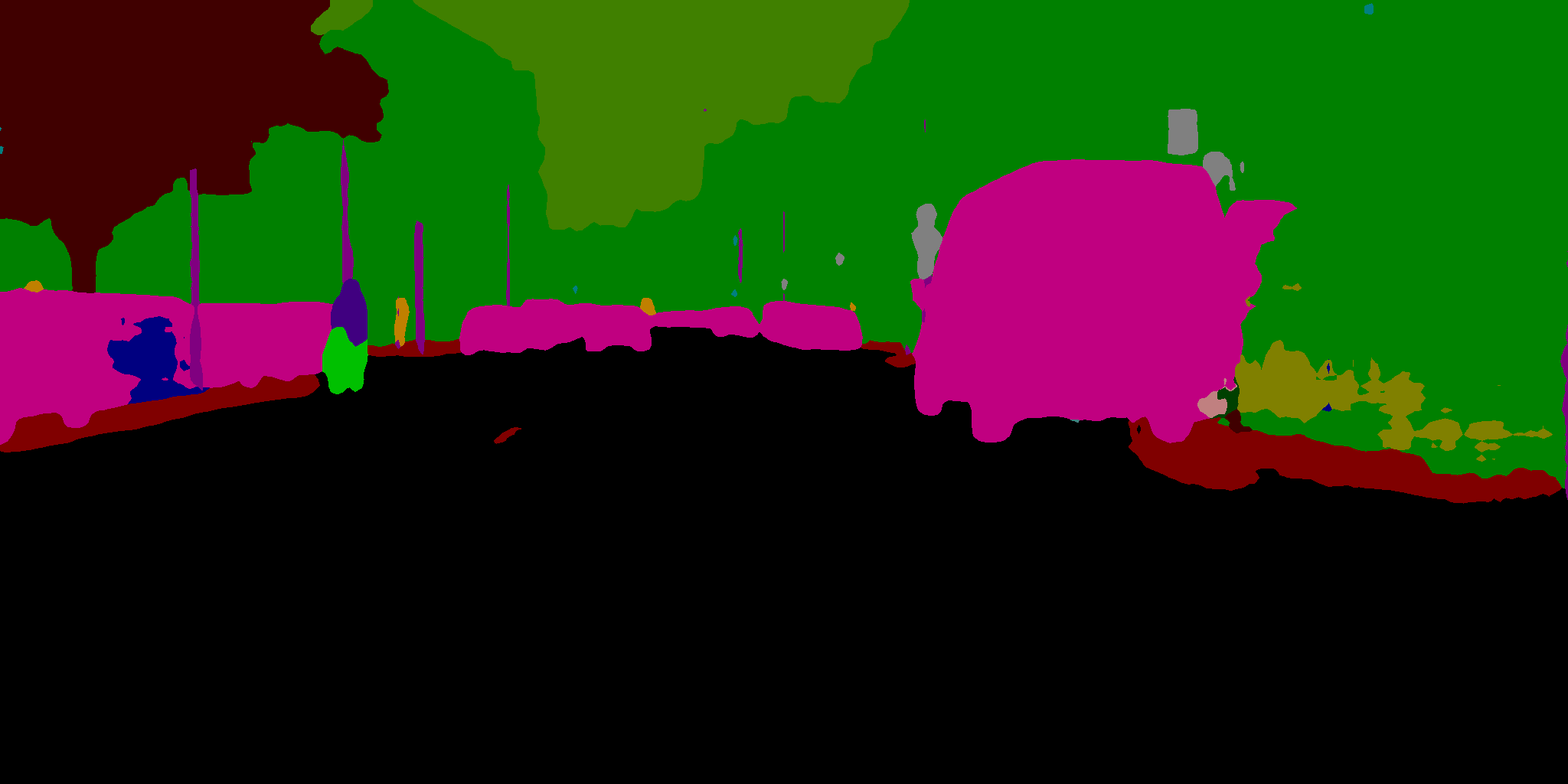}\label{fig:e5}}
\subfigure[\textit{BL 4}, Dice = $0.83$]{ \includegraphics[width=0.28\linewidth]{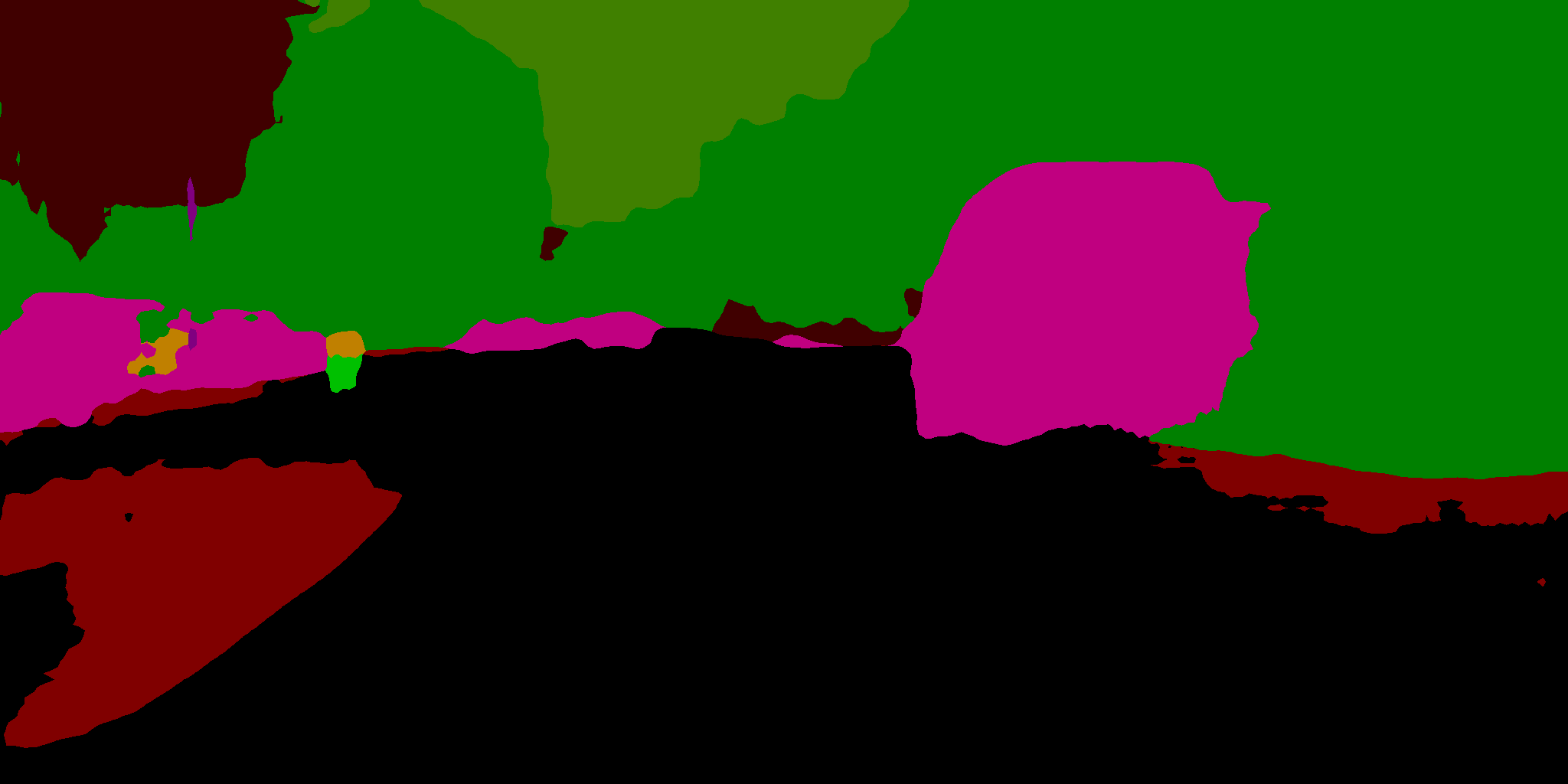}\label{fig:jpeg}}
\caption{Predicted segmentation mask for the $net_{seg,D}$ and baselines along with comparison with the ground truth.}
\label{fig:Seg_masks}
\end{figure*}

\subsection{Evaluation of Segmentation}
\noindent\textbf{Quantitative results:} The dice coefficient values for the baselines and $net_{seg, D}(\cdot)$, which is trained using compressed representations, are reported in Table~\ref{tab:net_seg_dice}. The results indicate that $net_{seg, D}(\cdot)$ performs similarly to \textit{BL 3} and \textit{BL 4} in terms of dice coefficient. This suggests that the compressed representations produced by $net_{C}(\cdot)$ contain significant semantic information that can be leveraged for other image analysis tasks, even though $net_{C}(\cdot)$ was not explicitly trained for this purpose.

\begin{table}[!h]
\caption{Quantitative and computational performance comparison with different baselines}
        \centering
        \begin{adjustbox}{max width=0.95\linewidth}
        \begin{tabular}{|c|c|c|c|c|c|c|c|c|c|}
            \hline
            Method & CF & Dice& SSIM & PSNR & Compute metric & $net_{C}(\cdot)$ & $net_{D}(\cdot)$ & $net_{seg,E}(\cdot)$ & $net_{seg,D(\cdot)}$ \\
            \hline
            \multirow{2}{*}{$net^{d=1}_{seg,D'}$} & \multirow{2}{*}{16} & \multirow{2}{*}{$0.88$} & \multirow{2}{*}{$0.99$} & \multirow{2}{*}{$45.74$} & Params($\times 10^{6}$) & $0.12$ & - & - & $53.41$
            \\
            \cline{6-10}
            & & & & & Flops & $16.40$ & - & - & $2053.53$ 
            \\
            \hline
            \multirow{2}{*}{$net^{d=2}_{seg,D'}$} & \multirow{2}{*}{66} & \multirow{2}{*}{$0.84$} & \multirow{2}{*}{$0.97$} & \multirow{2}{*}{$40.09$} & Params($\times 10^{6}$) & $0.19$ & - & - & $53.41$
            \\
            \cline{6-10}
            & & & & & Flops & $18.13$ & - & - & $513.41$ 
            \\
            \hline
            \multirow{2}{*}{$net^{d=3}_{seg,D'}$} & \multirow{2}{*}{230} & \multirow{2}{*}{$0.77$} & \multirow{2}{*}{$0.94$} & \multirow{2}{*}{$35.52$} & Params($\times 10^{6}$) & $0.27$ & - & - & $53.41$
            \\
            \cline{6-10}
            & & & & & Flops & $18.57$ & - & - & $125.93$ 
            \\
            \hline
            \multirow{2}{*}{BL 1} & \multirow{2}{*}{66} &\multirow{2}{*}{$0.91$} & - & - & Params($\times 10^{6}$) & - & - & $0.12$ & $53.40$
            \\
            \cline{6-10}
            & & & & & Flops($\times 10^{9}$) & - & - & $35.11$ & $513.30$ 
            \\
            \hline
            \multirow{2}{*}{BL 2} & \multirow{2}{*}{66} &\multirow{2}{*}{$0.90$} & - & - & Params($\times 10^{6}$) & $0.19$ & $0.31$  & $0.12$  & $53.41$
            \\
            \cline{6-10}
            & & & & & Flops($\times 10^{9}$) & $18.13$ & $9.48$ & $55.11$ & $513.40$
            \\
            \hline
            \multirow{2}{*}{BL 3} & \multirow{2}{*}{66} &\multirow{2}{*}{$0.88$} & - & - & Params($\times 10^{6}$) & $0.19$ & $0.31$  & $0.12$  & $53.41$
            \\
            \cline{6-10}
            & & & & & Flops($\times 10^{9}$) & $18.13$ & $9.48$ & $55.11$ & $513.40$
            \\
            \hline
            \multirow{2}{*}{BL 4} & \multirow{2}{*}{66} &\multirow{2}{*}{$0.82$} & \multirow{2}{*}{$0.76$} & \multirow{2}{*}{$25.07$} & Params($\times 10^{6}$) & - & -  & - & -
            \\
            \cline{6-10}
            & & & & & Flops($\times 10^{9}$) & - & -  & - & -
            \\
            \hline
        \end{tabular}
        \end{adjustbox}
        \label{tab:net_seg_dice}
\end{table}
%\FloatBarriers

Further, it can be observed that increasing the value of $d$, which results in a deeper network and higher compression factor, results in poorer reconstruction from the compressed representation owing to loss of information ~\cite{b5}. We observe a similar reduction in the quality of the segmentation map generated from compressed representations when identical decompressor architecture is used, as shown in Table ~\ref{tab:net_seg_dice}.

\noindent\textbf{Computational performance:} The total multiply-accumulate operations (MAC) to be performed and total trainable parameters that need to be tuned in order to perform segmentation using compressed representations received from an autonomous vehicle are shown in Table \ref{tab:net_seg_dice}. It is to be noted that for \textit{BL $2$} and \textit{BL $3$}, the cost of decompression also adds up to the total cost. It can be observed that the total computational cost for segmenting compressed representation is $11-12\%$ lower than that of \textit{BL $3$} and \textit{BL $1$}.

\noindent\textbf{Qualitative results:} The segmentation masks produced by $net_{seg,D}(\cdot)$ and baselines and the corresponding ground truth masks are shown in Fig.~\ref{fig:Seg_masks}.

\section{Conclusion}
We demonstrate that the compression codec proposed in ~\cite{b4, b5} is generalizable and can be adapted to other domains. Experimentally we achieved a compression factor of up to $66 \times$ without significant degradation in reconstruction quality. The compression factor can be further increased based on the application-specific threshold for acceptable reconstruction quality. We also prove that these compressed representations retain features beneficial for applications beyond compression without explicit training for the additional function. Experimentally we prove for the segmentation task where an average dice coefficient of $0.86$ was achieved for segmentation maps generated from representations compressed at a factor of $66$. In the future, we aim to extend this to other image analysis tasks like object detection and classification.

% \bibliographystyle{IEEEbib}
% {\footnotesize
% \bibliography{icme_template}}
% \end{document}

% --------------------------------------

\end{document}